\documentclass[twocolumn]{article}

\usepackage[width=17cm,height=22cm]{geometry}
\usepackage[utf8]{inputenc}
\usepackage{fancyvrb}
\usepackage{authblk}

\usepackage{graphicx} % more modern
\usepackage{subfigure} 

\usepackage[numbers]{natbib}

\usepackage{algorithm}
\usepackage{algorithmic}

\usepackage{amssymb}
\usepackage{amsmath}
\usepackage{graphicx}
\usepackage[utf8]{inputenc}
\usepackage{color}
\usepackage{algorithm}

\usepackage{url}
\usepackage{wasysym} % needed for some symbols varhexagon hexagon pentagon
\usepackage{xstring}
\usepackage{graphicx}
\usepackage{float}
\usepackage{rotating}
\usepackage{lscape}
\usepackage{colortbl}
\usepackage[table]{xcolor}
\usepackage{url}

\usepackage{amssymb,amsmath}
\usepackage{graphicx}
\usepackage{xcolor}
\usepackage{float}
\usepackage{rotating}

\newcommand{\includeperfprof}[1]{\includegraphics[width=0.5\textwidth,trim=0mm 0mm 0mm 10mm, clip]{#1}}	% As Is
\newcommand{\bbobdatapath}{ppdatamany/} % default output folder of rungeneric.py

\graphicspath{{\bbobdatapath}}

\def\RR{{\rm I\hspace{-0.50ex}R}}

\def\PrevName{ACM-ES}

\def\SAM{$^{s\ast}$}
\def\ALGOname{\SAM\hspace{-0.50ex}\PrevName}

\def\Fs{\mbox{$\widehat{f}$}}

\def\XX{KL-ACM-ES}
\newcommand{\vc}[1]{\textit{\textbf{#1}}}

\newcommand{\mstr}[1]{\mathrm{#1}}
\newcommand{\C}{ \ensuremath{\ma{C}} }

\def\Pt{\mbox{$P_{\theta_t}$}}
\def\P{\mbox{$P_{\theta}$}}
\def\Pp{\mbox{$P_{\theta'}$}}
\def\E{\mbox{${\cal E}_{\theta}$}}
\def\Ep{\mbox{${\cal E}_{\theta'}$}}
\def\F{\mbox{$\hat f_{\theta}$}}

\def\errth{\mbox{$\tau_{err}$}}

\title{KL-based Control of the Learning Schedule \\for Surrogate Black-Box Optimization}
\author[1]{Ilya Loshchilov}
\author[2]{Marc Schoenauer}
\author[3]{Mich\`ele Sebag}
\affil[1]{LIS, EPFL, Switzerland. ilya.loshchilov@epfl.ch}
\affil[2]{TAO, INRIA Saclay, France. marc.schoenauer@inria.fr}
\affil[3]{CNRS, LRI UMR 8623, France. michele.sebag@lri.fr}

\begin{document}
\date{}
\maketitle

\begin{abstract} 
This paper investigates the control of an ML component within the Covariance Matrix Adaptation Evolution Strategy (CMA-ES) devoted to black-box optimization. The known CMA-ES weakness is its sample complexity, the number of evaluations of the objective function needed to approximate the global optimum. This weakness is commonly addressed through surrogate optimization, learning an estimate of the objective function a.k.a. surrogate model, and replacing most evaluations of the true objective function with the (inexpensive) evaluation of the surrogate model. This paper presents a principled control of the learning schedule (when to relearn the surrogate model), based on the Kullback-Leibler divergence of the current search distribution and the training distribution of the former surrogate model. The experimental validation of the proposed approach shows significant performance gains on a comprehensive set of ill-conditioned benchmark problems, compared to the best state of the art including the quasi-Newton 
high-precision BFGS method.
\end{abstract}

\medskip

\noindent\textbf{Mots-clef}: expensive black-box optimization, evolutionary algorithms, surogate models, Kullback-Leibler divergence, CMA-ES.

%%%%%%%%%%%%%%%%%%%%%%%%%%%%%%%%%%%%%
\section{Introduction}\label{Introduction}
%%%%%%%%%%%%%%%%%%%%%%%%%%%%%%%%%%%%%
As noted in \cite{NIPSW12}, %(\textcolor[rgb]{1,0,0}{double-blind review?}), 
the requirements on machine learning algorithms (ML) might be rather different depending on whether these algorithms are used in isolation, or as components in computational systems. 
Beyond the usual ML criteria of consistency and convergence speed, % \cite{1995VapnikNaturalStatisticalTheory}, 
 ML components should enforce some stability and controllability properties. Specifically, an ML component should {\em never} cause any catastrophic event (be it related to exceedingly high computational cost or exceedingly bad performance), 
over all system calls to this component. As a concrete example of controllability-enforcing strategy and contrarily to the ML usage, the stopping criterion of a Support Vector Machine algorithm should be defined in terms of number of quadratic programming iterations, rather than in terms of accuracy, since the convergence to the global optimum might be very slow under some circumstances \cite{2009ListSVMlinesearch}. In counterpart, such a strategy might reduce the predictive performance of the learned model in some cases, thus hindering the performance of the whole computational system. It thus becomes advisable that the ML component takes in charge the control of its hyper-parameters and even the schedule of its calls (when and how the predictive model should most appropriately be rebuilt). The automatic hyper-parameter tuning of an ML algorithm in the general case 
however shows to be a critical task, requiring sufficient empirical evidence. 

This paper focuses on the embedding of a learning component within a distribution-based optimization 
algorithm \cite{2004RubinsteinCrossEntropyMethod,2003HansenCMA}. The visibility of such algorithms, particularly for industrial applications, is explained from their robustness w.r.t. (moderate) noise and multi-modality of the objective function \cite{2012HansenCMAESapplication}, in contrast to classical optimization methods such as quasi-Newton methods (e.g. BFGS \cite{1970ShannoBFGS}). One price to pay for this robustness
 is that the lack of any regularity assumption on the objective function leads to a large empirical {sample complexity}, i.e. number of evaluations of the objective function needed to approximate the global optima. 
Another drawback is the 
usually large number of hyper-parameters to be tuned for such algorithms to reach good performances. 
We shall however restrict ourselves in the remainder of the paper to the Covariance Matrix Adaptation Evolution Strategy (CMA-ES) \cite{2003HansenCMA}, known as an almost parameterless distribution-based black-box optimization algorithm. This property is commonly attributed to CMA-ES invariance properties w.r.t. both monotonous transformations of the objective function, and linear transformations of the initial representation of the instance space (section \ref{cma}).

The high sample complexity commonly prevents distribution-based optimization algorithms from being used on 
expensive optimization problems, where a single objective evaluation might require up to several hours (e.g.
for optimal design in numerical engineering). 
The so-called surrogate optimization algorithms (see \cite{2011JinAdvances} for a survey) address this limitation by coupling black-box optimization with learning of surrogate models, that is, local approximations of the objective function, and replacing most evaluations of the true objective function with the (inexpensive) evaluation of the surrogate function (section \ref{acm}). The key issue of surrogate-based optimization is the control of the learning module (hyper-parameters tuning and update schedule). 

In this paper, an integrated coupling of distribution-based optimization and rank-based learning algorithms,
called \XX, is presented. The contribution compared to the state of the art  \cite{2012LoshchilovSAACMGECCO} is to analyze the learning schedule with respect to the drift of the sample distribution: Formally, after the surrogate model is trained from a given sample distribution, this distribution is iteratively modified along optimization. When to relearn the surrogate model depends on how fast the error rate of the surrogate model increases as the sample distribution moves away from the training distribution. 
Under mild assumptions, it is shown that the error rate increase can be bounded with respect to the Kullback-Leibler divergence between the training and the current distribution, yielding a principled 
learning schedule. The merit of the approach is empirically demonstrated as it  shows significant performance gains on a comprehensive set of ill-conditioned benchmark problems \cite{2009HansenNoiselessTestbed,2009HansenNoisyTestbed}, compared to the best state of the art including the quasi-Newton high-precision BFGS method.

The paper is organized as follows. For the sake of self-containedness, section \ref{sec:soa} presents 
the Covariance Matrix Adaptation ES, and briefly reviews related work.  Section \ref{newmechanism} gives an overview of the proposed \XX\ algorithm and discusses the notion of drifting error rate. 
The experimental validation of the proposed approach is reported and discussed in section \ref{experimentalvalidation} and section \ref{conclusion} concludes the paper.

\def\Fst{\mbox{\Fs$_\alpha$}}
\def\err{{\mbox{Err}}}
\def\F{\Fst}
\section{State of the art}\label{sec:soa}
This section summarizes Covariance Matrix Adaptation Evolution Strategy (CMA-ES
), before discussing work related to surrogate-assisted optimization. 
% La ref Jin2011 est ds Swarm Computation... je ne la cite pas trop.
%the interested reader is referred to \cite{2011JinAdvances} for a more comprehensive presentation of surrogate distribution-based optimization.

\subsection{CMA-ES}\label{cma}
Let $f$ denote the objective function to be minimized on $\RR^n$:
\[ f: \RR^n \mapsto \RR \]
% XXX Ilya, a-t-on besoin du \mu_w plus loin ? 
The so-called $(\mu/ \mu_w,
\lambda)$-CMA-ES \cite{2003HansenCMA} maintains a Gaussian
distribution on $\RR^n$, iteratively used to generate $\lambda$ samples, and updated based on the best (in the sense of $f$) $\mu$ samples out of the $\lambda$ ones. Formally, samples $\vc{x}_{t+1}$ at time $t+1$ are drawn from the current Gaussian distribution $\Pt$, with $\theta_t = (\vc{m}_t, \sigma_t, \vc{C}_t)$: 
\begin{equation}
  \vc{x}_{t+1} \sim {\mathcal N}  \hspace{-0.13em}\left({\vc{m}_t,{\sigma_t}^2 \vc{C}_t}\right) %= \vc{m}_t + \sigma_t {\mathcal N} ( \vc{0}, \vc{C}_t),
  \end{equation}
		where $\vc{m}_t \in \RR^n$, $\sigma_t \in \RR$, and $\vc{C}_t \in \RR^{n \times n}$ respectively are the center of the Gaussian distribution (current best estimate of the optimum), the perturbation step size and  the covariance matrix.
	The next distribution center $\vc{m}_{t+1}$ is set to the weighted sum of the best $\mu$ samples%\footnote{Note that some active CMA-ES variants also use the worst samples to update $\vc{m}_{t+1}$.}
, denoting $\vc{x}_{t+1}^{(i:\lambda)}$ the $i$-th best sample out of the $\lambda$ ones:
%XXX On peut peut-etre enlever la footnote.
	\begin{equation}
  \vc{m}_{t+1} = \sum_{i=1}^{\mu} w_i \vc{x}_{t+1}^{(i:\lambda)} \mbox{~with~} \sum_{i=1}^{\mu} w_i=1
\end{equation}	
The next covariance matrix $\vc{C}_{t+1}$ is updated from $\vc{C}_{t}$ using both the local information 
about the search direction, given by 
$\frac{1}{\sigma_t} 
(\vc{x}_{t+1}^{(i:\lambda)} - \vc{m}_t)$, and the global information stored in the so-called evolution path $\vc{p}_{t+1}$ of the distribution center $\vc{m}$. 
For positive learning rates $c_1$ and $c_{\mu}$ ($c_1 + c_{\mu} \leq 1$) 
the update of the covariance matrix reads:
\begin{equation}
\begin{array}{rl}
 \lefteqn{ \vc{C}_{t+1} = (1-c_1-c_{\mu})\vc{C}_t + c_1 \underbrace{\vc{p}_{t+1} \cdot {\vc{p}_{t+1}}^{T}}_{\mstr{rank-one\,update}} }\\
	& + c_{\mu} \underbrace{ \sum^{\mu}_{i=1} \frac{w_i}{{\sigma_t}^2} \, (\vc{x}_{t+1}^{(i:\lambda)} - \vc{m}_t) \cdot (\vc{x}_{t+1}^{(i:\lambda)} - \vc{m}_t)^{T}}_{\mstr{rank-\mu \,update}}
\end{array}
\end{equation}
The step-size $\sigma_{t+1}$ is likewise updated to best align the distribution 
of the actual evolution path of $\sigma$, and an evolution path under random selection.
\def\C{\mbox{$\vc{C}$}}

As mentioned, the CMA-ES robustness and performances \cite{2009HansenBBOBcomparing31algo} are explained from its invariance properties. On the one hand, CMA-ES only considers the sample ranks after $f$; it thus does not make any difference between optimizing $f$ and $g \circ f$, for any strictly increasing scalar function $g$. 
%As will be seen, this property sharply constrasts with e.g., the behavior of quasi-Newton methods. 
On the other hand, the self-adaptation of the covariance matrix \C\ makes CMA-ES invariant w.r.t.
orthogonal transformations of the search space (rotation, symmetries, translation).
%\footnote{If the initial candidate solutions are transformed accordingly.}, due to the adaptation of the covariance matrix $\vc{C}$. 
% 		\item \textcolor[rgb]{1,0,0}{\textbf{THIS CAN BE REMOVED IF NEEDED}} Scale invariance if the initial scaling, e.g., $\sigma^0$, and the initial search point(s) are chosen accordingly.
% 	\end{itemize}
Interestingly, CMA-ES can be interpreted in the Information-Geometric Optimization (IGO) framework \cite{2011OllivierIGO}. IGO achieves a \textit{natural gradient ascent} on the space of parametric distributions on the sample space $X$, using the Kullback-Leibler divergence as distance among distributions. It has been shown that the basic $(\mu,\lambda)$-CMA-ES variant is a particular case of IGO when the parametric distribution space is that of Gaussian distributions \cite{2011OllivierIGO}.

\def\Fst{\mbox{\Fs}}
\def\nh{\mbox{$\hat n$}}
\subsection{Surrogate-based CMA-ES}\label{saACM}\label{acm}
\begin{figure}[t]
\centerline{ 
	\includegraphics[width=3.4truein]{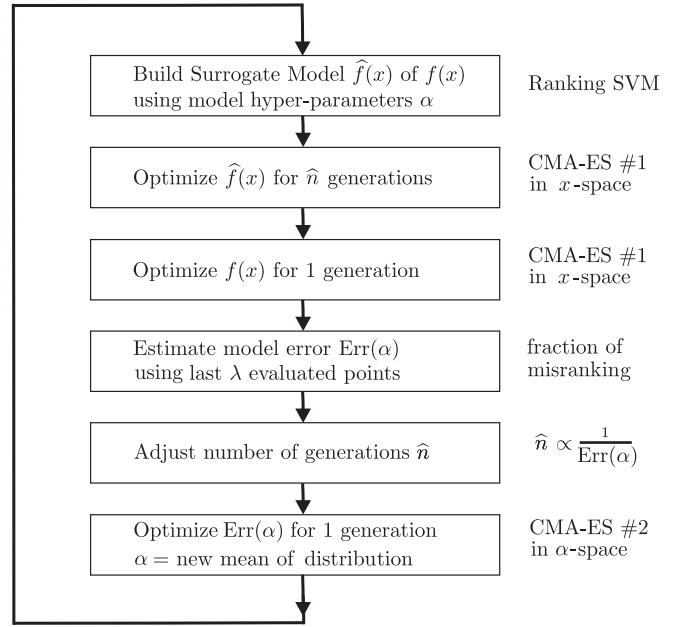}
}
\caption{\ALGOname: interleaved optimization loop. %If use active versions of the IPOP-CMA-ES with restarts (IPOP-aCMA-ES) as CMA-ES \#1 and \#2, two surrogate-assisted versions can be recovered: IPOP-saACM-ES and IPOP-aACM-ES with and without model hyper-parameters optimization respectively. 
}
\label{figscheme}
\end{figure}

%XX Ilya, je garde Buche parce que c'est important de mentionner Gaussian Processes pour les reviewers ML.
Since the late 90s, many learning algorithms have been used within surrogate-based optimization, ranging 
from neural nets to Gaussian Processes (a.k.a. kriging) \cite{2003UlmerESandGP,2005BucheGPOP}, using in particular the expected improvement \cite{1998JonesEGO} as selection criterion. 
The surrogate CMA-ES algorithm most similar to our approach, \ALGOname\ \cite{2012LoshchilovSAACMGECCO}, interleaves two optimization procedures (Fig. \ref{figscheme}):\\
The first one (noted CMA-ES \#1) regards the optimization of the objective function $f$, assisted by the surrogate model \Fst; the second one (CMA-ES \#2) regards the optimization of the learning hyper-parameters $\alpha$ used to learn \Fst.  

%In both cases, the original \textit{generation/iteration} procedure of CMA-ES is not modified, but called when necessary in order to change the state of CMA-ES in the search space, that also leads to a change of the internal parameters of CMA-ES (i.e., $\vc{m}$, $\vc{C}$, $\sigma$ and evolution paths for $\sigma$ and $\vc{m}$).

More precisely, \ALGOname\ first launches CMA-ES on the true objective function $f$ for 
% Je mets n_{start} par consistence avec \hat n
a number of iterations $n_{start}$, and gathers the computed samples in a training set \E\ $= \{(x_i,f(x_i)), i = 1, \ldots q \}$; the first surrogate model \F\ is learned from \E. 
\ALGOname\ then iterates the following process, referred to as epoch:\\
i) \F\ is
optimized by CMA-ES for a given number of steps \nh, leading from distribution \P\ to \Pp;\\ ii)  
CMA-ES is launched with distribution \Pp\ on the true objective function, thereby building a new training set \Ep;\\ iii) the previous and current training sets \E\ and \Ep\ are used to adjust \nh\ and the other learning hyper-parameters $\alpha$, and a new surrogate model \F\ is learned from \Ep. 

\subsubsection*{Surrogate model learning}
Surrogate model \Fst\ is learned from the current training set \E, using Ranking-SVM 
\cite{2005JoachimsSVMperf} together with the learning hyper-parameter vector $\alpha$. For the sake of computational efficiency, a linear number of ranking constraints ($x_i \prec x_{i+1},~ i=1 \ldots q-1$) is used (assuming wlog that the samples in \E\ are ordered by increasing value of $f$). By construction, the surrogate model thus is invariant under monotonous transformations of $f$. 

The invariance of \Fst\ w.r.t. orthogonal transformations of the search space is 
enforced by using a Radial Basis Function (RBF) kernel, which involves the inverse of the covariance matrix adapted by CMA-ES. Formally, the rank-based surrogate model is learned using the kernel $K_C$ defined as:  $K_C(\vc{x}_i,\vc{x}_j) = \mstr{e}^{-\frac{(\vc{x}_i-\vc{x}_j)^T\vc{C}^{-1}(\vc{x}_i-\vc{x}_j)}{2 s^2}}$, 
which corresponds to rescaling datasets \E\ and \Ep\ using the transformation
$\vc{x} \rightarrow \vc{C}^{-1/2}(\vc{x} - \vc{m})$, where $\vc{C}$ is the covariance matrix adapted by CMA-ES (and $s$ is a learning hyper-parameter). Through this kernel, \Fst\ benefits from the CMA-ES efforts in identifying the local curvature of the optimization landscape. 

\def\DER{drift error rate}

\subsubsection*{Learning hyper-parameters}
As mentioned, the control of the learning module (e.g. when to refresh the surrogate model; how many ranking 
constraints to use; which penalization weight) has a dramatic impact on the performance of the surrogate 
optimization algorithm. Both issues are settled in \ALGOname\ through exploiting the error of the 
surrogate model \Fst\ trained from \E, measured on \Ep. 
Formally, let $\ell(\F,x,x')$ be 1 iff \F\ misranks $x$ and $x'$ compared to $f$, 1/2 if $f(x) = f(x')$ and 0 otherwise. The empirical ranking error of \Fst\ on \Ep, referred to as empirical \DER, is defined as\\

\centerline{$\displaystyle \widehat{Err} = \frac{1}{|\Ep|(|\Ep|-1)}\sum_{\stackrel{x, x' \in \Ep} { x \neq x'}} \ell(\F, x, x')$ }

It is used to linearly adjust \nh: close to 50\%, it indicates that \Fst\ is no better than random guessing and that the surrogate model should have been rebuilt earlier (\nh = 0); close to 0\%, it inversely suggests that  \Fst\ could have been used for a longer epoch (\nh\ is set to $n_{max}$, user-specified parameter of \ALGOname). 

In the same spirit, \E\ and \Ep\ are used to adjust the learning hyper-parameter vector, through a 
1-iteration CMA-ES on the hyper-parameter space, minimizing the \DER\ of the surrogate model learned from \E\ with hyper-parameters $\alpha'$. The learning hyper-parameter vector $\alpha$ to be used in the next epoch is set to the center of the CMA-ES distribution on the hyper-parameter space. 

\subsection{Discussion}
The main strength of \ALGOname\ is to achieve the simultaneous optimization of the objective function 
$f$ together with the learning hyper-parameters (considering \nh\ as one among the learning hyper-parameters), thus only requiring the user to initially define their range of values and adjusting them 
online to minimize the \DER. It is worth noting that the automatic tuning of ML
hyper-parameters is critical in general, particularly so when dealing with small 
sample sizes. The fact that the hyper-parameter tuning was found to be effective in the considered setting seems to be explained as the ML component in \ALGOname\ i) actually considers a sequence of learning problems
defined by the successive distributions \P, ii) receives some feedback in each epoch about the $\alpha$ choice made in the previous epoch.

A significant weakness however is that the \ALGOname\ learning schedule is defined in terms of 
the number \nh\ of CMA-ES iterations in each epoch. However, the \DER\ should rather depend on how fast the optimization distribution \P\ 
is modified. This remark is at the core of the proposed algorithm.

% \section{KL}
% Lemma 361\footnote{{\tt http://www.stat.cmu.edu/~cshalizi/754/2006/} Lecture 28, p 192.} (Divergence and Total Variation): For any two distributions P and Q, 
% $$D_{KL}(P||Q) \geq \frac{1}{2 \ln 2} ||P-Q||^2_1$$
% {\bf Proof}: Algebra. See, e.g., Cover and Thomas (1991, Lemma 12.6.1, pp. 300–301).

\section{Kullback-Leibler Divergence for Surrogate Model Control}
\label{newmechanism}
\def\adm{\mbox{$Err_{admissible}$}}
\def\R{\mbox{$Err_{\P}(\F)$}}
\def\Rm{\mbox{$\widehat{Err_{\P,q}(\F)}$}}
\def\Rpm{\mbox{$\widehat{Err_{\Pp,q}(\F)}$}}
\def\Rp{\mbox{$Err_{\Pp}(\F)$}}
\def\Fh{\mbox{$\widehat{f_\theta}$}}
\def\Fhp{\mbox{$\hat f'$}}
\def\KL{\mbox{$KL_{thresh}$}}
\def\KK{\mbox{$KL(\Pp || \P)$}}

\def\q{q}  % la quintessence de la macro

The proposed \XX\ algorithm presented in this section differs from 
\ALGOname\ regarding the control of the learning schedule, 
that is, the decision of relearning the surrogate model. The proposed criterion is based on the Kullback-Leibler divergence between the distribution \P\ that was used to generate the training set of the current surrogate model \F, referred to as training distribution,
\ %(\textcolor[rgb]{1,0,0}{pas training, mais de CMA}) ?? pas compris
and the current working distribution \Pp\ of CMA-ES.  It is worth noting that since the Kullback-Leibler divergence depends only on \P\ and not on the parameterization of $\theta$, this criterion is intrinsic and could be used for other distribution-based black-box optimization algorithms. 

\subsection{Analysis}
Letting \P\ and \Pp\ denote two distributions on the sample space, their Kullback-Leibler divergence noted
\KK\ is defined as:
% Let \P\ denote in the following the distribution on the sample space used to train the 
% surrogate model noted \Fh. Let \Pp\ denote the current distribution on the sample space. 
% Let the Kullback Leibler divergence between distributions \P\ and \Pp\ be defined as:
\begin{equation}
\label{eq:KL}
\KK = \int_{x} \ln \frac{\Pp(x)}{\P(x)} \Pp(dx).
\end{equation}

Let \F\ denote in the following the surrogate model learned from \E, sampled after distribution \P. 
%The idea is to bound the generalization and empirical ranking error of \Fh\ as the sample distribution drifts from \P\ to \Pp.
The idea is 
to retrain the surrogate model \F\ whenever it is estimated that its ranking error on \Pp\ might be greater 
than a (user-specified) admissible error \adm. Let us show that the difference between the generalization 
error of \F\ wrt \P\ and \Pp\ is bounded depending on the KL divergence of \P\ and \Pp:

\noindent{\bf Proposition 1}\\
The difference between the expectation of the ranking error of \F\ w.r.t. \P\ and w.r.t. \Pp, respectively noted \R\ and \Rp, is bounded by the square root of the KL divergence of \P\ and \Pp:
\begin{equation}
 |\R - \Rp| \leq c_k \sqrt{KL (\Pp || \P)}
\label{eq:1}
\end{equation}
with $c_k = 2\sqrt{2\ln{2}}$

{\em Proof}
\[ \begin{array}{lll}
    |\R - \Rp| \\
\;\;\; =  |\iint \ell(\F,x,x') [\P(x)\P(x') - \Pp(x)\Pp(x')] dx dx' |\\
\;\;\; \leq |\iint (\P(x)\P(x') - \Pp(x)\Pp(x')) dx dx'|  \\
\;\;\; \leq  2 ||\P - \Pp||_1\\
\;\;\;  \leq  c_k \sqrt{KL (\Pp || \P)}
   \end{array}
\]
where the first inequality follows from the fact that $\ell$ is positive and bounded by 1, 
and the second from applying the usual trick $(ab-cd) = a(b-d) + d(a-c)$, and separately integrating
w.r.t. $x$ and $x'$.
% \[ (\P(x)\P(x') - \Pp(x)\Pp(x'))= \P(x)(\P(x') - \Pp(x')) + \Pp(x')(\P(x) - \Pp(x))\]
% therefore
% \[ |\iint (\P(x)\P(x') - \Pp(x)\Pp(x')) dx dx'| < ||\P - \Pp||_1 \] 
The last inequality follows from the result that for any two distributions $P$ and $Q$, 
\[ KL(P||Q) \geq \frac{1}{2 \ln 2} ||P-Q||^2_1 \]
See, e.g., Cover and Thomas (1991, Lemma 12.6.1, pp. 300–301). \hfill$\Box$

The second step, bounding the difference between the empirical and the generalization ranking error of \Fh,  follows from the statistical learning theory applied to ranking 
\cite{2008ClemencconRanking,2009AgarwalGeneralization,2012RejchelRanking}.

Let us first recall the property of {\em uniform loss stability}. A ranking algorithm has the uniform loss 
stability property iff for any  two $q$-size samples ${\cal E}$ and ${\cal E}'$  drawn after the same distribution and differing by a single sample, if \F\ and \Fhp\ are the ranking functions learned from respectively ${\cal E}$ and ${\cal E}'$, the following holds for any $(x,x')$ pair, for some  $\beta_q$ that only depends on $q$:
\[ |\ell(\F,x,x') - \ell(\Fhp,x,x')| < \beta_q \]

\noindent{\bf Proposition 2} \cite{2009AgarwalGeneralization}\\
If the ranking algorithm has the uniform loss stability property, then for any $0 < \delta < 1$, with probability at least $1 - \delta$ (over the draw of the $q$-size test set \E\ drawn according to \P), 
the generalization ranking error of \F\ is bounded by its empirical error on \E, plus a term that only depends on $\q$: 
\begin{equation}
 |\R - \Rm| <  2\beta_q+(q\beta_q+1)\sqrt{\frac{2 \ln{1/\delta}}{q}}
\label{eq:2}
\end{equation}

From Eqs. (\ref{eq:1}) and (\ref{eq:2}), it is straightforward to show that with probability at least $1 - \delta/2$ the empirical error of \F\ on \Ep\ is bounded by a term that only depends on $\q$, plus the square root of the KL divergence between \P\ and \Pp:
\begin{equation}
\begin{array}{ll}
\medskip \Rpm  < 4\beta_q+2(q\beta_q+1)\sqrt{\frac{2 \ln{1/\delta}}{q}}\\
\;\;\;\;\;\; + \Rm + c_k \sqrt{KL(\P||\Pp)} 
\end{array}
\label{equ:generalisation}
\end{equation}

In the context of this work, Ranking-SVM have the uniform loss stability property \cite{2009AgarwalGeneralization}. Hence, if the $q$-dependent terms of Eq. (\ref{equ:generalisation}) are small enough, it is possible, given a parameter \adm, to define a threshold \KL\ such that, provided that the KL divergence between \P\ and \Pp\ remains below  \KL, the empirical error of \Fh\ on \Ep\ remains bounded by \adm\ with high probability. 
In practice however, as noted by \cite{2008ClemencconRanking}, it is well known that the above bound 
is often quite loose, even in the case where $\ell$ is convex and $\sigma$-admissible \cite{2009AgarwalGeneralization}. For instance, using the hinge loss function proposed in \cite{2009AgarwalGeneralization}, the constant $\beta_q$ of Eq. (\ref{equ:generalisation}) is inversely proportional to $\q$, and the $\q$-dependent terms of Eq. (\ref{equ:generalisation}) vanish for large $\q$. However, as discussed in section \ref{Introduction}, the context of application of ML algorithms is driven here by the optimization goal. In particular, the number of training samples $\q$ has to be kept as small as possible. This is why an empirical alternative for the update of \KL\ has been investigated, inspired by the trust-region paradigm from classical optimization.

\subsection{Adaptive adjustment of \KL}
Let us remind that, for \P\ and \Pp\ (respectively defined by ${\theta}=(\vc{m}, \vc{C})$ and $\theta'=(\vc{m}', \vc{C'})$) the Kullback-Leibler divergence of \P\ and \Pp\ has a closed form expression,
 
\begin{equation}
\begin{array}{rl}
\label{eq:gaussKL}
\lefteqn{ KL(P_{\theta'} || P_{\theta}) = \frac{1}{2} [ tr(\vc{C}^{-1} \vc{C'}) +} \\
& (\vc{m}-\vc{m}')^T \vc{C}^{-1}(\vc{m} - \vc{m}') - n - \ln(\frac{det \vc{C'}}{det \vc{C}}) ].
\end{array}
\end{equation}
\begin{figure}[tb]
\begin{center}
  \includegraphics[width=0.49\textwidth]{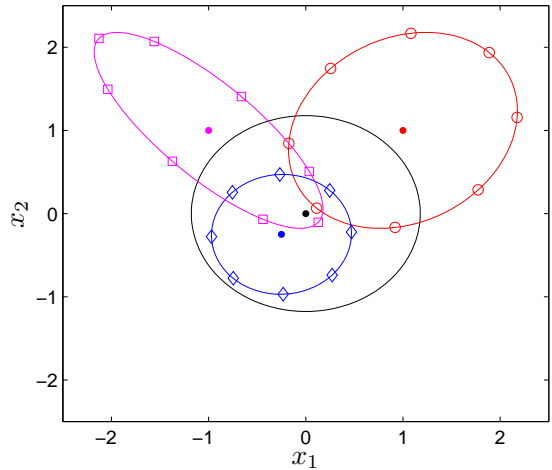}  %\includegraphics[scale=0.34]{pprldmany_20D_noiselessall}
\end{center}
\caption{\label{fig:KL} Ellipsoidal 95\% confidence regions of the Gaussian distribution \P\ (black thin line) and three other Gaussian distributions \Pp\  (color marked lines) with same \KK.}
\end{figure}

Note that increasing values of \KK\ might reflect different phenomenons; they can be due to either differences in the distribution center or in the scaling of the covariance matrix (see the examples on Fig. \ref{fig:KL}). 

The sought threshold \KL\ finally is interpreted in terms of trust regions \cite{1983MoreTrustRegionStep}. Classical heuristic optimization methods often proceed by associating to a region of the search space 
the (usually) quadratic surrogate model approximating the objective function in this region. Such a 
region, referred to as trust region, is assessed from the ratio of expected improvement measured on the surrogate model, and the improvement on the true objective. Depending on this ratio, the trust region is 
expanded or restricted.

The proposed KL-based control of the learning schedule can be viewed as a principled way to adaptively
control the dynamics of the trust region, with three differences. Firstly, the trust region 
is here defined in terms of distributions on the search space: the trust region is defined as the set of 
all distributions \Pp\ such that \KK $<$ \KL. Secondly, this trust region is assessed a posteriori 
from the empirical error of the surrogate model, on the first distribution \Pp\ {\em outside} the trust region. Thirdly, this assessment is exploited to adjust the KL ``radius'' of the next trust 
region, \KL.

Finally, \KL\ is set such that $\log{\KL}$ is inversely proportional to the relaxed surrogate error $Err$: 
\begin{equation}
\label{eq:errfunc}
\ln(\KL) \leftarrow  \frac{\errth - Err}{\errth} \ln(KL_{Max}),
\end{equation}
where \errth\ is an error threshold (experimentally set to .45), ${KL}_{Max}$ is the maximal allowed KL divergence when \F\ is ideal, and the relaxed surrogate error $Err$ is computed as $ Err = (1 - \alpha) Err + \alpha \Rpm $, in order 
to moderate the effects of \Rpm\ high variance. 
The log-scaling is chosen on the basis of preliminary experiments; its adjustment online is left for further work.
% Note that log-scaling is chosen here as a simple approximation, while an optimal scaling indeed is problem-dependent and should be adapted during the search.

Finally, \XX\ differs from \ALGOname\ in two points:\\
$\bullet$ The second  step of the algorithm (Fig. \ref{figscheme}) thus becomes "Optimize \F\ while $\KK \leq \KL$; \\
$\bullet$ The fifth  step of the algorithm (Fig. \ref{figscheme})  becomes "Adjust $\KL$" from Eq. (\ref{eq:errfunc}).

\section{Experimental validation}\label{experimentalvalidation}

\def\pbd{pBFGS**2}
\def\pbq{pBFGS**4}

\begin{figure*}[tb]
\begin{center}
  \includegraphics[width=0.6\textwidth]{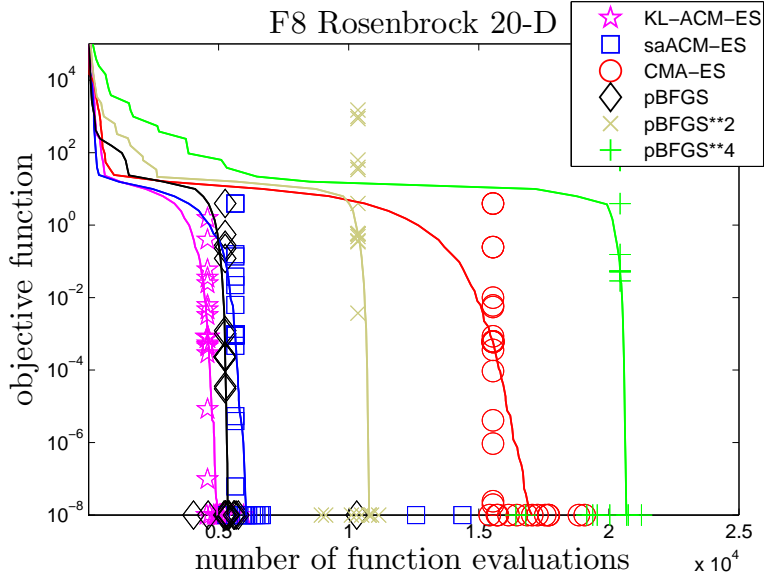}  %\includegraphics[scale=0.34]{pprldmany_20D_noiselessall}
\end{center}
\caption{\label{fig:pBFGSnoiselessRos} Comparative performance of \XX\ compared to high-precision BFGS and 
CMA-ES variants (see text) on the 20-dimensional Rosenbrock function $f_{Ros}$, $f_{Ros}^2$ and $f_{Ros}^4$.
 The medium number of function evaluations (out of 15 runs) to reach the target objective value $10^{-8}$ was computed and the corresponding run is shown. 
Markers show the objective value reached in each run after a given number of function evaluations.}
\end{figure*}

This section presents the experimental validation of the proposed \XX\ algorithm compared to various CMA-ES algorithms including \ALGOname, and the quasi Newton BFGS algorithm \cite{1970ShannoBFGS} on the BBOB 
noiseless and noisy benchmark suite \cite{2012HansenBBOBsetup}. 

%%%%%%%%%%%%%%%%%%%%%%%%%%%%%%%%%%%%%
\subsection{Experimental Setting}
%%%%%%%%%%%%%%%%%%%%%%%%%%%%%%%%%%%%%
For reproducibility, the Matlab source code of \XX\ is made available together with its default parameters\footnote{https://sites.google.com/site/acmesKL/}. After preliminary experiments, $\ln(KL_{Max})$ is set to 6; the CMA-ES $\lambda$ parameter used when optimizing \Fh\ is multiplied by 100 when the empirical error rate of \Fh\ on \Ep\ is less than .35. The comparative validation firstly involves \ALGOname\ with its default parameters\footnote{https://sites.google.com/site/acmesgecco/}. The CMA-ES variant used within \ALGOname\ and \XX\ is the state of the art BIPOP-active CMA-ES algorithm \cite{2010HansenBBOBActiveCMA}.
The comparative validation also involves the Quasi-Newton BFGS method. Indeed, BFGS suffers from known numerical problems on ill-conditioned problems (see 
\cite{1987PowellBFGS} for an extensive discussion). This limitation is overcome by considering instead the 32-decimal digit precision  arithmetic version of BFGS, referred to as pBFGS\footnote{https://sites.google.com/site/highprecisionbfgs/. \\ For gradient approximations by finite differences $\epsilon=10^{-20}$ is used in pBFGS instead of $\epsilon=10^{-8}$ in BFGS.} and included in the high-precision arithmetic package ARPREC \cite{2002BaileyARPREC}. 

The algorithms have been compared on the twenty-four 20-dimensional noiseless 
and thirty noisy benchmark problems of the BBOB suite \cite{2009HansenNoiselessTestbed,2009HansenNoisyTestbed} with different known characteristics: 
	separable, non-separable, unimodal, multi-modal, ill-conditioned, deceptive, functions with and without weak global structure. For each problem, 50 uniformly generated orthogonal transformations of $f$ are considered. 

For each problem and each algorithm, 15 independent runs (each one with a randomly chosen position for the optimum) are launched. The performance of 
each algorithm is reported as the median optimum value (in log scale) vs the number of evaluations of $f$ (Fig. \ref{fig:pBFGSnoiselessRos}), or as the empirical cumulative distribution of success for solving sets of similar functions (Fig. \ref{fig:ECDFs20D}, see caption there). All algorithms are initialized with samples uniformly drawn in $[-5,5]^{20}$.

\def\fros{\mbox{$f_{Ros}$}}

\subsection{Case study: the Rosenbrock function}
A first study is conducted on the 20-dimensional Rosenbrock function and its second and fourth power. Rosenbrock function is defined as:
\[ \fros({x})=\sum_{i=1}^{n-1}\left(100 (x^{2}_{i}-x_{i+1})^{2}+(x_{i}-1)^{2}\right) \]

The empirical results (Fig. \ref{fig:pBFGSnoiselessRos}) show that both surrogate-based optimization
algorithms \ALGOname\ and \XX\ improve on the CMA-ES variant baseline by a factor ranging from 2.5 to 3. 

\XX\ outperforms \ALGOname\ and pBFGS on the Rosenbrock function (BFGS, omitted for clarity, behaves like pBFGS as the Rosenbrock function is not ill-conditioned). The merits of the invariance w.r.t. monotonous transformations of the objective function shine 
as the pBFGS behavior is significantly degraded on $f_{Ros}^2$ (legend \pbd) and $f_{Ros}^4$
(legend \pbq) compared to \fros, slowing down the convergence by a factor of about 4 for $f_{Ros}^4$. 
Quite the contrary, all CMA-ES variants including \ALGOname\ and \XX\ behave exactly the same on all three functions, by construction.

The performance improvement of \XX\ on \mbox{\ALGOname} and pBFGS is over 20\%. 
%The KL-based \ALGOname\ outperforms the original \ALGOname, partially due to a larger population size used to optimize $\hat{f}$.

\subsection{Results on BBOB problems}

\begin{figure*}
 \begin{tabular}{@{}c@{}c@{}}
 separable fcts & moderate fcts \\
 \includeperfprof{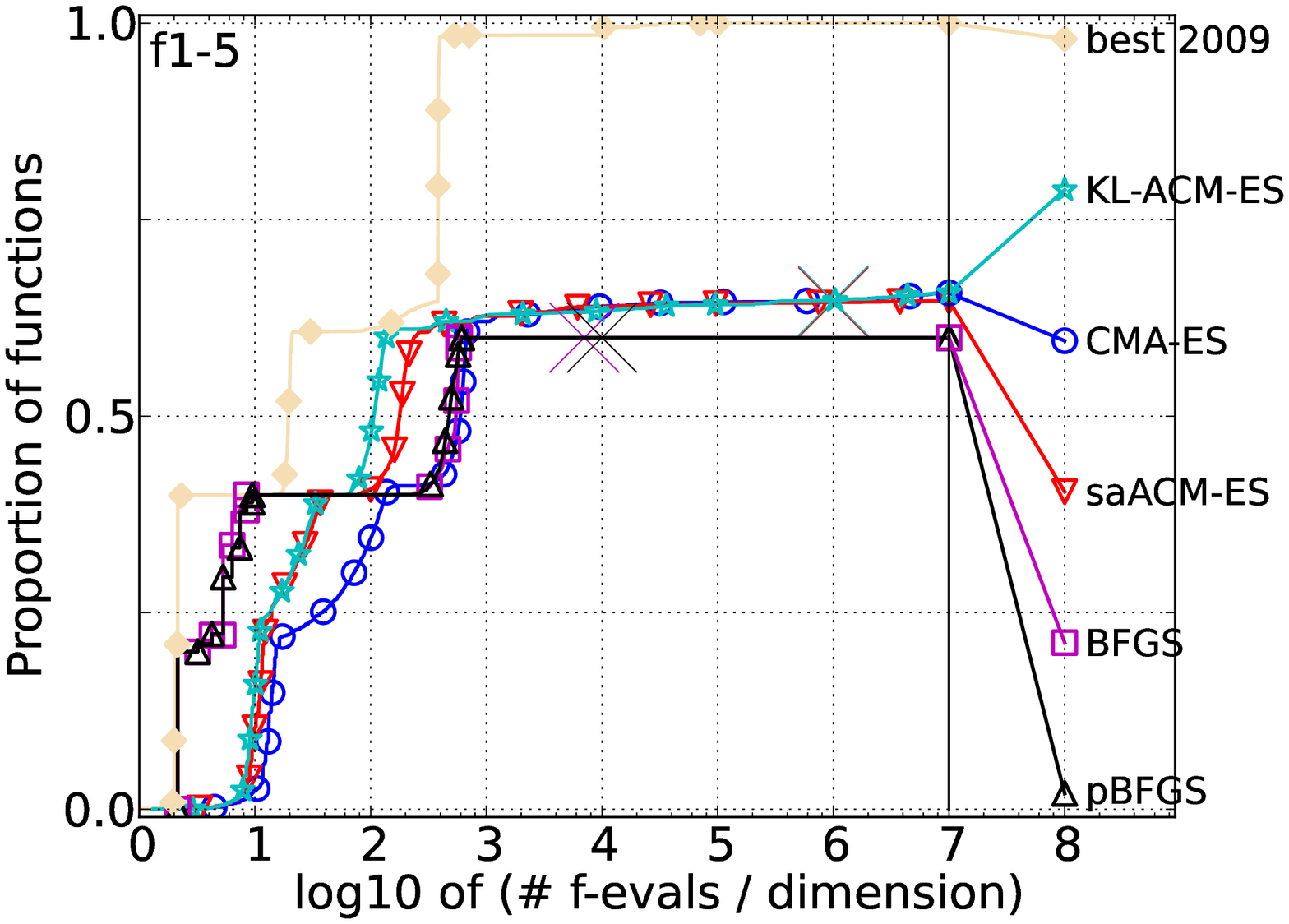} &
 \includeperfprof{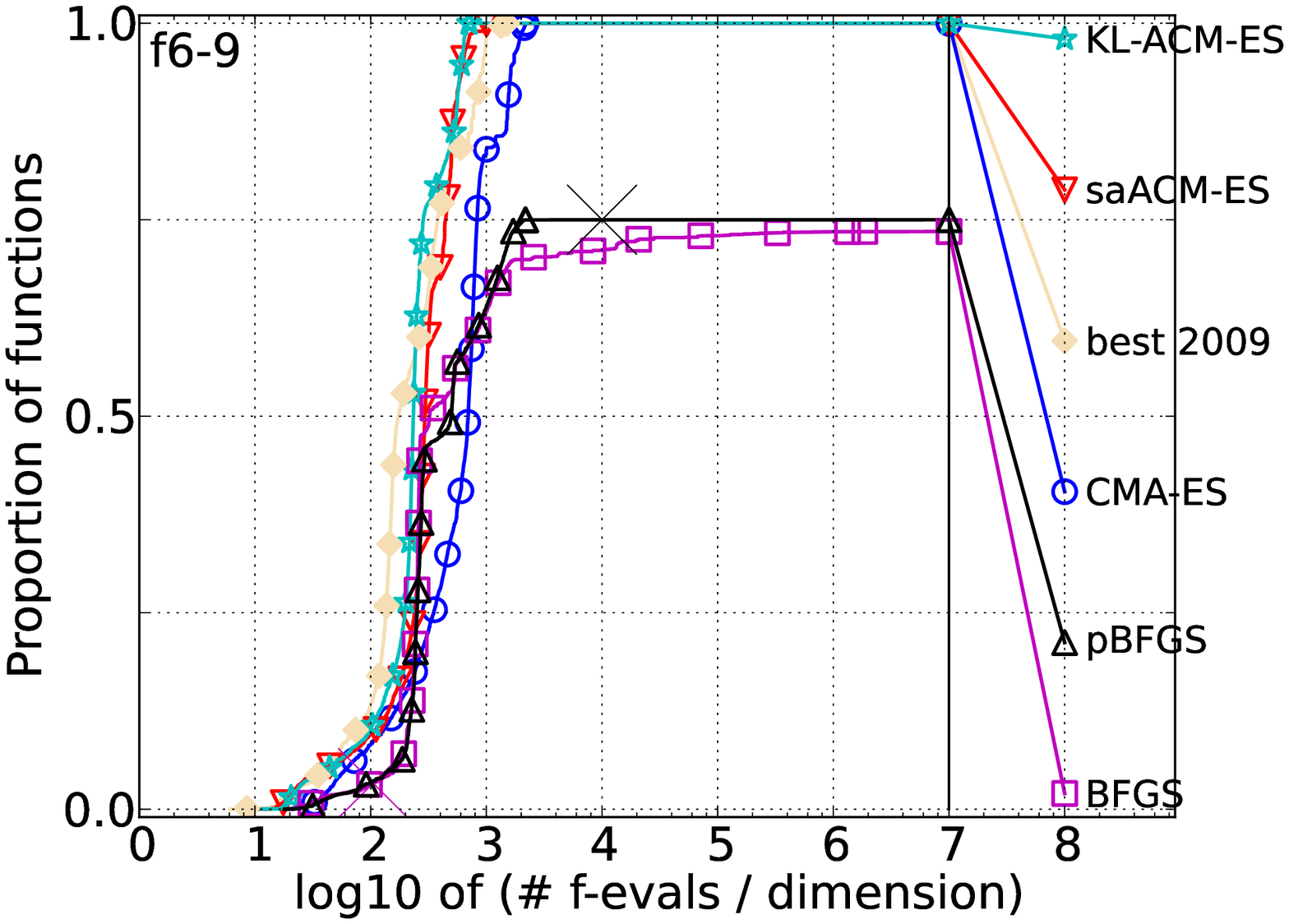} \\ 
ill-conditioned fcts & multimodal fcts \\
 \includeperfprof{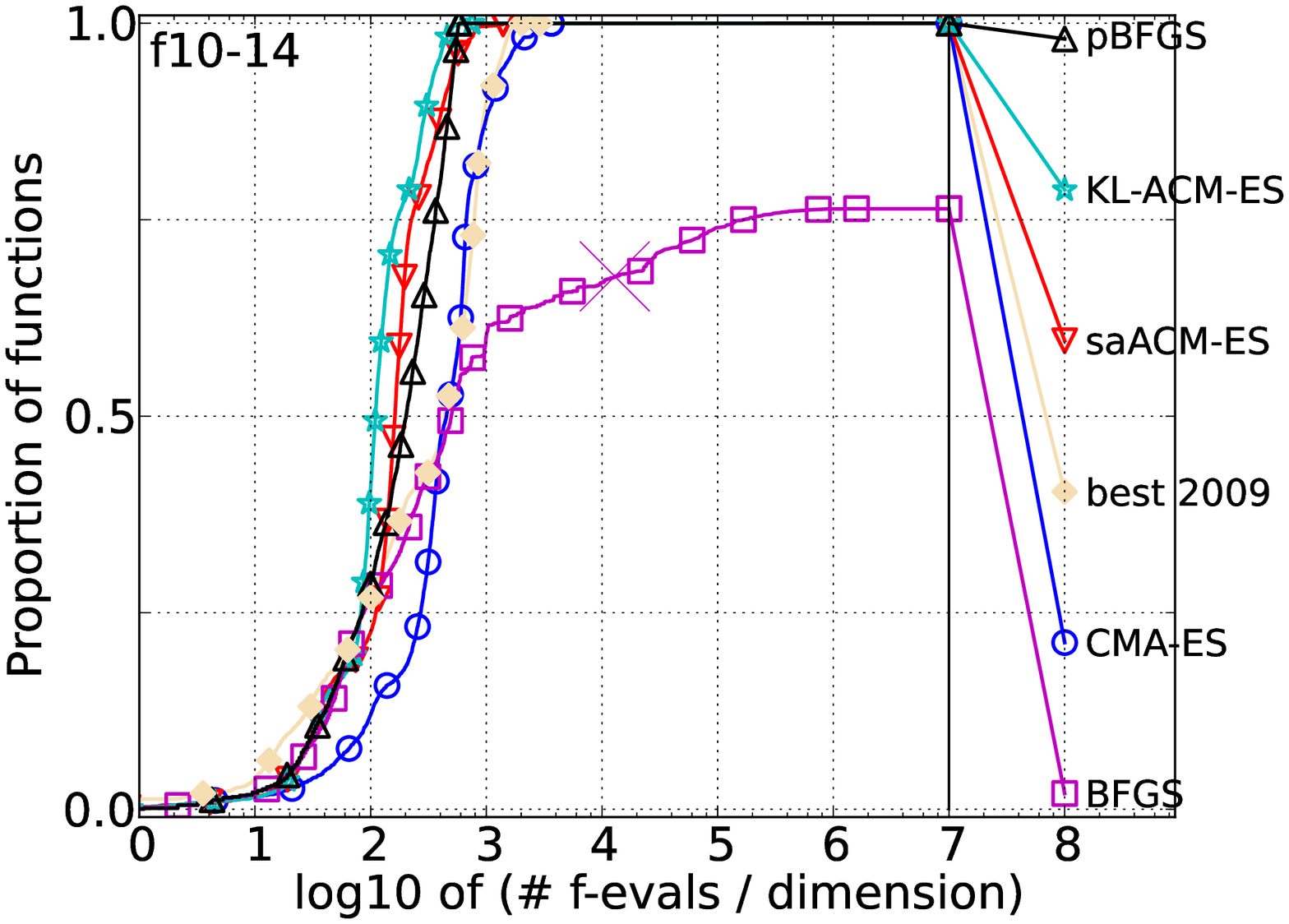} &
 \includeperfprof{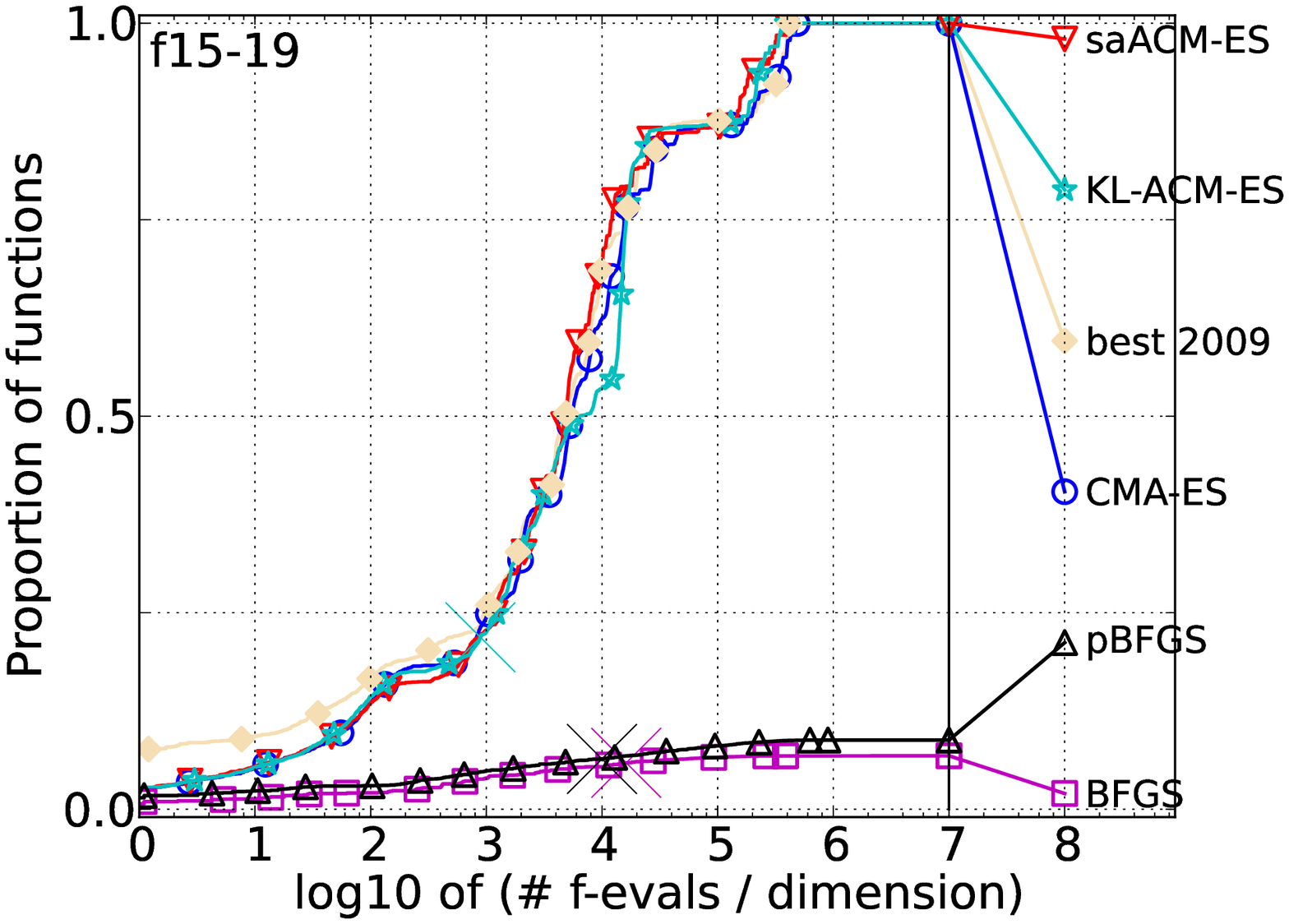} \\ 
 weakly structured multimodal fcts & all functions\\
 \includeperfprof{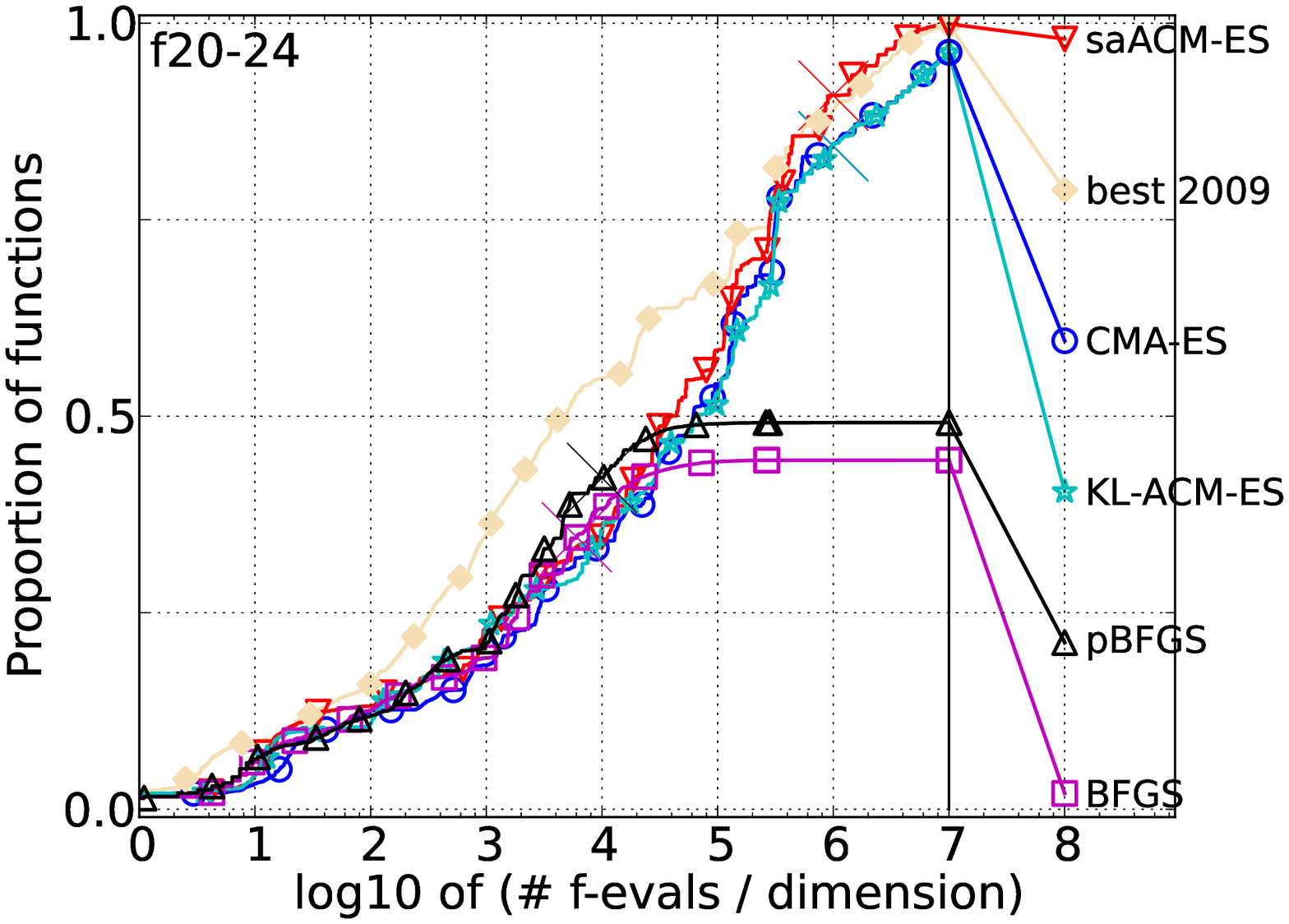} & 
 \includeperfprof{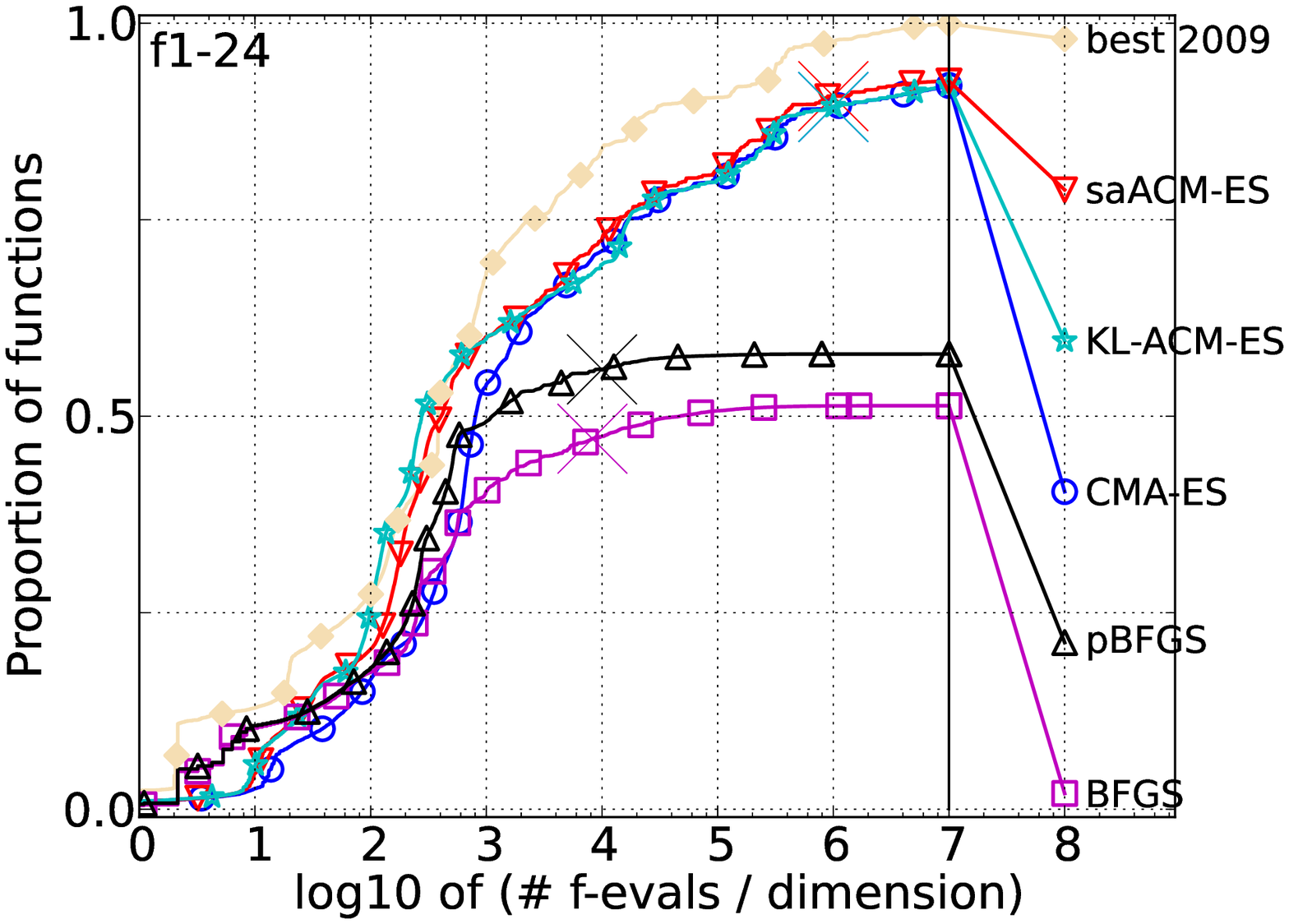} 
 \end{tabular}
\caption{
\label{fig:ECDFs20D}
Empirical cumulative distributions (i.e. proportion of functions solved -- up to given target precisions in $10^{[-8..2]}$) of the number of objective function evaluations
divided by dimension for all functions (last plot) and subgroups of similar functions (other plots) in 20-D.  The "best 2009" line indicates the BBOB 2009 ``portfolio oracle``, the aggregation of the best results for each function. 
The proposed algorithm is depicted as \XX. A detailed description of these representations of BBOB results can be found in \cite{2009HansenBBOBcomparing31algo}.
}
\end{figure*}

Fig. \ref{fig:ECDFs20D} displays the overall results on BBOB benchmark. Similar functions are grouped, thus distinguishing the cases of separable, moderately difficult, ill-conditioned, 
multi-modal, weakly structured multi-modal objective functions (last plot aggregates all functions).
The {\em Best 2009} reference corresponds to the (virtual) best performance reached over the portfolio of all algorithms participating in the BBOB 2009 contest (portfolio oracle; note that the high-precision pBFGS was not included in the portfolio). 

On separable objective functions, pBFGS dominates all CMA-ES variants for small numbers of evaluations. Then \XX\ catches up, followed by \ALGOname\ and the other CMA-ES variants. On moderate and ill-conditioned functions, \XX\ dominates all other algorithms and it even improves over {\em Best 2009}. On multi-modal and weakly structured 
multi-modal functions, \XX\ is dominated by \ALGOname; in the meanwhile BFGS and pBFGS alike yield poor performance. 

Finally, on the overall plot, \XX\ shows good performances compared to the state of the art, prodding the {\em Best 2009} 
curve in the median region due to the significant progress made on ill-conditioned functions.

Besides the merits of rank-based optimization, these results demonstrate the relevance of using 
high-precision BFGS instead of BFGS for solving ill-conditioned optimization problems. 
A caveat is that high-precision computations require the source code to be rewritten (not only in the optimization algorithm, but also in the objective function), which is hardly feasible in standard black-box scenarios $-$ even when the objective source code is available.\\
Albeit pBFGS significantly dominates BFGS, its behavior is shown to be significantly
degraded when the scaling of the objective function differs from the "desirable" quadratic BFGS scaling.

\section{Discussion and future work}
\label{conclusion}

This paper investigates the control of an ML component within a compound computational system. 
In order to provide a steady and robust contribution to the whole system, it is emphasized that an ML component must take in charge the adjustment of its learning hyper-parameters, and also {\em 
when and how frequently this ML component should be called}. 
In the meanwhile and as could have been expected, such an autonomous ML component must accommodate applicative priorities and experimental conditions which differ from those commonly faced by ML 
algorithms in isolation.

A first attempt toward building such an autonomous ML component in the context of black-box distribution-based optimization has been presented. In this context, the ML component is meant to supply an estimate of the objective optimization function; it faces a sequence of learning problems, drawn from moving distributions on the instance space. The first contribution is to show how the 
KL divergence between successive distributions can yield a principled control of the learning schedule, that is, the decision to relearn an estimate of the objective optimization function. 
The second contribution is that the \XX\ algorithm implementing this learning schedule control improves 
on the best state-of-the-art distribution-based optimization algorithms and quasi-Newton methods, 
on a comprehensive suite of ill-conditioned benchmark problems. 

Further work will examine how to further enhance the autonomy of the ML component, e.g. when facing multi-modal objective functions.  Alternative comparison-based surrogate 
models will also be considered, such as  Gaussian Processes for ordinal regression \cite{2005ChuPreferenceGP}. Finally, as shown by \cite{2012HennigQuasiNewton}, quasi-Newton methods can be interpreted as approximations of Bayesian linear regression under varying prior assumptions;
a prospective research direction is to replace the linear regression by ordinal regression-based Ranking SVM or Gaussian Processes in order to derive a version of BFGS invariant w.r.t. monotonous 
transformations of the objective function $f$. 

%%%%%%%%%%%%%%%%%%%%%%%%%%%%%%%%%%%%%
%\section{Acknowledgments}
%%%%%%%%%%%%%%%%%%%%%%%%%%%%%%%%%%%%%

%.

%.

%.
%This work was partially funded by FUI of System@tic Paris-Region ICT cluster through contract DGT 117 407 {\em Complex Systems Design Lab} (CSDL). 

\bibliography{mybib}

\newcommand{\etalchar}[1]{$^{#1}$}
\begin{thebibliography}{HFRA09b}

\bibitem[AAHO11]{2011OllivierIGO}
L.~{Arnold}, A.~{Auger}, N.~{Hansen}, and Y.~{Ollivier}.
\newblock {Information-Geometric Optimization Algorithms: A Unifying Picture
  via Invariance Principles}.
\newblock {\em ArXiv e-prints}, June 2011.

\bibitem[AN09]{2009AgarwalGeneralization}
S.~Agarwal and P.~Niyogi.
\newblock {Generalization Bounds for Ranking Algorithms via Algorithmic
  Stability}.
\newblock {\em J. of Machine Learning Research}, 10:441--474, 2009.

\bibitem[BSK05]{2005BucheGPOP}
D.~Buche, N.~N. Schraudolph, and P.~Koumoutsakos.
\newblock {Accelerating Evolutionary Algorithms with Gaussian Process Fitness
  Fn Models}.
\newblock {\em IEEE Trans. Systems, Man and Cybernetics}, 35(2):183--194, 2005.

\bibitem[BYLT02]{2002BaileyARPREC}
D.H. Bailey, H.~Yozo, X.S. Li, and B.~Thompson.
\newblock {ARPREC: An Arbitrary Precision Computation Package}.
\newblock Technical report, Ernest Orlando Lawrence Berkeley National
  Laboratory, CA, 2002.

\bibitem[CG05]{2005ChuPreferenceGP}
W.~Chu and Z.~Ghahramani.
\newblock {Preference learning with Gaussian processes}.
\newblock In {\em Proc. 22nd ICML}, pages 137--144. ACM, 2005.

\bibitem[CLV08]{2008ClemencconRanking}
S.~Clemen{\c{c}}on, G.~Lugosi, and N.~Vayatis.
\newblock {Ranking and Empirical Minimization of U-statistics}.
\newblock {\em The Annals of Statistics}, 36(2):844--874, 2008.

\bibitem[HAFR12]{2012HansenBBOBsetup}
N.~Hansen, A.~Auger, S.~Finck, and R.~Ros.
\newblock {Real-Parameter Black-Box Optimization Benchmarking 2012:
  Experimental Setup}.
\newblock Technical report, INRIA, 2012.

\bibitem[Han13]{2012HansenCMAESapplication}
N.~Hansen.
\newblock {References to CMA-ES Applications}.
\newblock Website, January 2013.
\newblock Available online at http://www.lri.fr/~hansen/cmaapplications.pdf.

\bibitem[HAR{\etalchar{+}}10]{2009HansenBBOBcomparing31algo}
N.~Hansen, A.~Auger, R.~Ros, S.~Finck, and P.~Po\v{s}\'{\i}k.
\newblock {Comparing results of 31 algorithms from the black-box optimization
  benchmarking BBOB-2009}.
\newblock In {\em GECCO Companion}, pages 1689--1696, 2010.

\bibitem[HCO12]{NIPSW12}
P.~Hennig, J.~P. Cunningham, and M.~A. Osborne, editors.
\newblock {\em Probabilistic Numerics Workshop, NIPS}, 2012.

\bibitem[HFRA09a]{2009HansenNoiselessTestbed}
N.~Hansen, S.~Finck, R.~Ros, and A.~Auger.
\newblock {Real-Parameter Black-Box Optimization Benchmarking 2009: Noiseless
  Functions Definitions}.
\newblock Research Report RR-6829, INRIA, 2009.

\bibitem[HFRA09b]{2009HansenNoisyTestbed}
N.~Hansen, S.~Finck, R.~Ros, and A.~Auger.
\newblock {Real-Parameter Black-Box Optimization Benchmarking 2009: Noisy
  Functions Definitions}.
\newblock Research Report RR-6869, INRIA, 2009.

\bibitem[HK12]{2012HennigQuasiNewton}
P.~Hennig and M.~Kiefel.
\newblock {Quasi-Newton Methods: A New Direction}.
\newblock {\em Preprint arXiv:1206.4602}, 2012.

\bibitem[HMK03]{2003HansenCMA}
N.~Hansen, S.D. M{\"u}ller, and P.~Koumoutsakos.
\newblock {Reducing the Time Complexity of the Derandomized Evolution Strategy
  with Covariance Matrix Adaptation}.
\newblock {\em Evolutionary Computation}, 11(1):1--18, 2003.

\bibitem[HR10]{2010HansenBBOBActiveCMA}
N.~Hansen and R.~Ros.
\newblock {Benchmarking a weighted negative covariance matrix update on the
  BBOB-2010 noiseless testbed}.
\newblock In {\em GECCO Companion}, pages 1673--1680, New York, NY, USA, 2010.
  ACM.

\bibitem[Jin11]{2011JinAdvances}
Y.~Jin.
\newblock {Surrogate-Assisted Evolutionary Computation: Recent Advances and
  Future Challenges}.
\newblock {\em Swarm and Evolutionary Computation}, pages 61--70, 2011.

\bibitem[Joa05]{2005JoachimsSVMperf}
T.~Joachims.
\newblock {A Support Vector Method for Multivariate Performance Measures}.
\newblock In {\em Proc. 22nd ICML}, pages 377--384. ACM, 2005.

\bibitem[JSW98]{1998JonesEGO}
Donald~R. Jones, Matthias Schonlau, and William~J. Welch.
\newblock {Efficient Global Optimization of Expensive Black-Box Functions}.
\newblock {\em J. of Global Optimization}, 13(4):455--492, December 1998.

\bibitem[LS09]{2009ListSVMlinesearch}
N.~List and H.U. Simon.
\newblock {SVM-optimization and steepest-descent line search}.
\newblock In {\em Proceedings of the 22nd COLT}, 2009.

\bibitem[LSS12]{2012LoshchilovSAACMGECCO}
I.~Loshchilov, M.~Schoenauer, and M.~Sebag.
\newblock {S}elf-{A}daptive {S}urrogate-{A}ssisted {CMA-ES}.
\newblock In T.~Soule and J.H. Moore, editors, {\em Proc. GECCO}, pages
  321--328. ACM Press, July 2012.

\bibitem[MS83]{1983MoreTrustRegionStep}
J.~J. Mor{\'e} and D.~C. Sorensen.
\newblock Computing a trust region step.
\newblock {\em SIAM Journal on Scientific and Statistical Computing},
  4(3):553--572, 1983.

\bibitem[Pow87]{1987PowellBFGS}
M.J.D. Powell.
\newblock {Updating Conjugate Directions by the BFGS Formula}.
\newblock {\em Mathematical Programming}, 38(1):29--46, 1987.

\bibitem[Rej12]{2012RejchelRanking}
Wojciech Rejchel.
\newblock On ranking and generalization bounds.
\newblock {\em The Journal of Machine Learning Research}, 98888:1373--1392,
  2012.

\bibitem[RK04]{2004RubinsteinCrossEntropyMethod}
R.~Y. Rubinstein and D.~P. Kroese.
\newblock {\em The {Cross-Entropy} Method: A Unified Approach to Combinatorial
  Optimization, {Monte-Carlo} Simulation and Machine Learning}.
\newblock Springer, 2004.

\bibitem[Sha70]{1970ShannoBFGS}
D.~F. Shanno.
\newblock {Conditioning of {Quasi-Newton} Methods for Function Minimization}.
\newblock {\em Mathematics of Computation}, 24(111):647--656, 1970.

\bibitem[USZ03]{2003UlmerESandGP}
H.~Ulmer, F.~Streichert, and A.~Zell.
\newblock {Evolution Strategies assisted by Gaussian Processes with Improved
  Pre-Selection Criterion}.
\newblock In {\em IEEE Congress on Evolutionary Computation}, pages 692--699,
  2003.

\end{thebibliography}

\end{document}